\documentclass[10pt,twocolumn,letterpaper]{article}

\usepackage{cvpr}              
\usepackage[accsupp]{axessibility} 

\usepackage{amsmath,amsfonts,bm}

\def\eqref#1{equation~\ref{#1}}

\def\1{\bm{1}}

\DeclareMathAlphabet{\mathsfit}{\encodingdefault}{\sfdefault}{m}{sl}
\SetMathAlphabet{\mathsfit}{bold}{\encodingdefault}{\sfdefault}{bx}{n}

\usepackage[dvipsnames]{xcolor}

\definecolor{cvprblue}{rgb}{0.21,0.49,0.74}
\usepackage[pagebackref,breaklinks,colorlinks,citecolor=cvprblue]{hyperref}

\def\paperID{7411} 
\def\confName{CVPR}
\def\confYear{2024}

\title{Image-Text Co-Decomposition for Text-Supervised Semantic Segmentation}

\author{Ji-Jia Wu$^{1}$ \qquad Andy Chia-Hao Chang$^{2}$ \qquad Chieh-Yu Chuang$^{2}$ \qquad Chun-Pei Chen$^{2}$ \qquad Yu-Lun Liu$^{2}$ \\ [2pt]
Min-Hung Chen$^{3}$ \qquad Hou-Ning Hu$^{4}$ \qquad Yung-Yu Chuang$^{1}$ \qquad Yen-Yu Lin$^{2}$ \\
\\ [2pt]
$^{1}$National Taiwan University~
$^{2}$National Yang Ming Chiao Tung University\\
$^{3}$NVIDIA~
$^{4}$MediaTek~
}

\usepackage{amsfonts}  

\newcommand{\ouralignmentpair}{region-word}
\newcommand{\ouralignment}{\ouralignmentpair~alignment}

\newcommand{\methodfull}{Image-Text Co-Decomposition}
\newcommand{\methodfullbold}{Image-Text \textbf{Co}-\textbf{De}composition}
\newcommand{\methodabbr}{CoDe}

\newcommand\CODE{\small\url{https://github.com/072jiajia/image-text-co-decomposition}}

\begin{document}
\maketitle
\begin{abstract}

This paper addresses text-supervised semantic segmentation, aiming to learn a model capable of segmenting arbitrary visual concepts within images by using only image-text pairs without dense annotations. 
Existing methods have demonstrated that contrastive learning on image-text pairs effectively aligns visual segments with the meanings of texts.
We notice that there is a discrepancy between text alignment and semantic segmentation: A text often consists of multiple semantic concepts, whereas semantic segmentation strives to create semantically homogeneous segments.
To address this issue, we propose a novel framework, \methodfullbold~(\textbf{\methodabbr}),
where the paired image and text are jointly decomposed into a set of image regions and a set of word segments, respectively, and contrastive learning is developed to enforce region-word alignment.
To work with a vision-language model, we present a prompt learning mechanism that derives an extra representation to highlight an image segment or a word segment of interest, with which more effective features can be extracted from that segment.
Comprehensive experimental results demonstrate that our method performs favorably against existing text-supervised semantic segmentation methods on six benchmark datasets.
The code is available at \CODE.

\end{abstract}
\section{Introduction}
\label{sec:intro}

\begin{figure*}[t]
    \centering
    \includegraphics[width=0.96\linewidth]{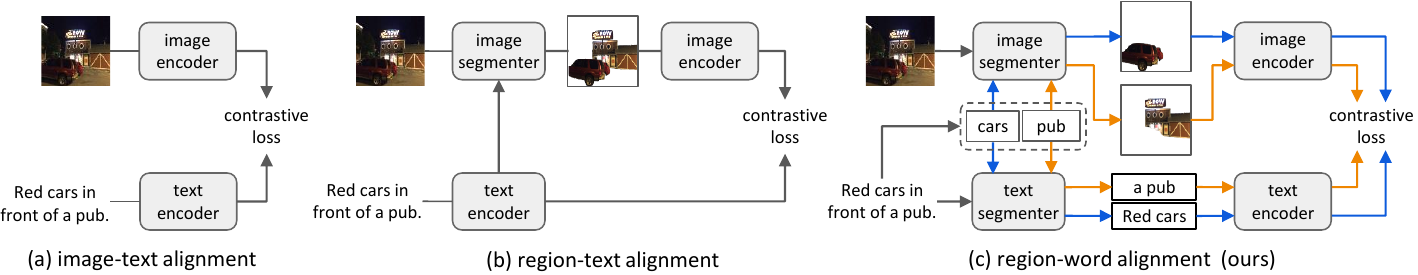}
    \vspace{-0.05in}
    \caption{\small
    Existing methods perform text-supervised semantic segmentation by learning either (a) image-text alignment or (b) region-text alignment.
    This paper presents (c) region-word alignment via image-text co-decomposition, where the image and the text are decomposed into object regions and word segments, respectively, while contrastive learning is used to establish cross-modal correspondences between these image and word segments.
    }
    \vspace{-0.15in}
    \label{fig:teaser}
\end{figure*}

Semantic segmentation is essential to various applications \cite{feng2020deep, diakogiannis2020resunet, yuan2021review} in computer vision but is hindered by several critical challenges.
First, the expensive cost of acquiring pixel-level annotations limits the applicability of fully supervised semantic segmentation methods. 
Second, most existing methods \cite{Strudel_2021_ICCV, Zheng_2021_CVPR, xie2021segformer} are developed to work on predefined categories and leave themselves inapplicable to rare or unseen concepts described by free-form text. 
To address these obstacles, a new research direction has emerged in vision-language models, referred to as {\em text-supervised semantic segmentation}~\cite{xu2022groupvit, liu2022vilseg, xu2023learning, xing2023rewrite, yi2023simple, cha2022tcl}. 
This task develops segmentation models capable of assigning labels across large vocabularies of concepts and supporting semantic segmentation model training without pixel-wise annotations.

\cref{fig:teaser} compares existing methods for text-supervised semantic segmentation by grouping their cross-domain alignment mechanisms into three categories, including {\em image-text}, {\em region-text}, and {\em \ouralignmentpair} alignment.
Despite the differences, most of these methods compensate for the lack of pixel-wise annotations on broad semantic concepts by exploring abundant image-text pairs on the internet.
The textual descriptions bring extensive knowledge across diverse categories.
Thus, existing methods typically apply a vision-language model such as CLIP~\cite{radford2021clip} to textual descriptions to acquire the semantic context of the corresponding images for segmentation model learning. 

The image-text alignment is widely adopted in the literature \eg~\cite{xu2022groupvit, liu2022vilseg, xing2023rewrite}. 
As depicted in \cref{fig:teaser}a, methods of this group derive an image encoder and a text encoder by aligning them in a joint embedding space.
They then use their proposed zero-shot transfer techniques to enable the two encoders to predict segmentation output.
Despite the simplicity, they introduce unfavorable discrepancies between the training and testing phases since we aim to match the semantic features from the text to the corresponding image segments rather than the whole image during testing.

To mitigate this issue, the region-text alignment is explored.
As shown in \cref{fig:teaser}b, methods of this group such as \cite{cha2022tcl} utilize a pre-trained visual-language model to derive an additional image segmenter that discovers concepts described by the text.
They enforce the consistency between the segmented region and the text but suffer from the discrepancy between the region-text alignment and semantic segmentation:
A text may consist of multiple concepts, such as {\em pub}, {\em night}, and {\em car} in \cref{fig:teaser}b, while semantic segmentation aims to identify regions of the same concept.

To address the aforementioned issues in the image-text and region-text alignments, we propose a novel framework, \methodfullbold~(\textbf{\methodabbr}), to achieve {\em \ouralignment}.
As illustrated in \cref{fig:teaser}c, we utilize a visual-language model to construct an image segmenter and a text segmenter: 
The former decomposes an image into image segments, while the latter decomposes a text into word segments. 
In addition, there exist one-to-one correspondence between image and word segments.
This way, the discrepancy between training and testing is alleviated since each image segment is derived from a single concept given by the corresponding word segment.

The proposed \methodabbr~framework comprises four components: an image segmenter, a text segmenter, a region-word alignment module, and a prompt learning module.
We randomly select nouns in the text.
For each selected noun \eg, ``car'', the image segmenter identifies the image segment that matches the noun, \ie, the region of the car, while the text segmenter discovers the corresponding word segment, \ie, ``red cars.''
The region-word alignment is developed to enforce the consensus between the image and word segments.
To better work with a vision-language model, we present a prompt learning module to derive an extra representation, enabling more effective feature extraction.

The main contributions of this work are as follows:
\begin{itemize}
\item We propose a new framework, \methodfull~(\methodabbr), to learn the \ouralignment~for eliminating train-test and image-text discrepancies, facilitating text-supervised semantic segmentation.
\item We propose a prompt learning method to address domain shift issues arising from blank areas during the highlighting process and enhance the alignment between highlighted regions and highlighted words.
\item Our method effectively carries out zero-shot semantic segmentation and performs favorably against the state-of-the-art methods on six benchmark datasets.
\end{itemize}

\section{Related Works}
\label{sec:related_works}

\subsection{Open-Vocabulary Semantic Segmentation}

Open-vocabulary semantic segmentation focuses on segmenting any concepts within images, even those unseen during training, based solely on textual descriptions. 
Its three important branches are discussed as follows:

\vspace{-0.18in}
\paragraph{Semi-supervised setting with mask-annotations.} Methods of this group such as \cite{li2022languagedriven, liang2023open, ghiasi2022scaling, 2022fusioner, Xu_2023_CVPR, Han_2023_ICCV} learn from dense annotations to produce high-quality segmentation masks, and then utilize image-text pairs and pre-trained vision-language models to extend the segmentation capability to a larger target vocabulary.
Despite the remarkable results, these methods are hindered by their reliance on costly dense annotations, posing a challenge in cases where such annotations are difficult to obtain.

\vspace{-0.18in}
\paragraph{Training-free methods.} Another line of research \eg \cite{zhou2022maskclip, shin2022reco, wang2023diffusion} makes the most of large pre-trained models for open-vocabulary segmentation without training.
MaskCLIP~\cite{zhou2022maskclip} introduces a modification to the final layer of the CLIP image encoder, yielding dense feature maps that could be employed as initial segmentation maps for further refinement. 
ReCo~\cite{shin2022reco} constructs an image archive and makes use of retrieval and co-segmentation to identify co-occurrence regions among a specific category. 
Although these methods eliminate the process of training, 
the results exhibit significant room for improvement, which shows the need for additional supervision to accomplish this task.

\vspace{-0.18in}
\paragraph{Text-supervised semantic segmentation.} It strikes a balance between the two aforementioned branches. 
Methods of this group are discussed in detail in the following because our method belongs to this group. 

\subsection{Text-Supervised Semantic Segmentation}
Text-supervised semantic segmentation \cite{xu2022groupvit, liu2022vilseg, xu2023learning, xing2023rewrite, yi2023simple, cha2022tcl, ren2023viewco, cai2023mixreorg, wu2023diffumask, pandey2023language} decomposes an image into semantic regions according to text descriptions.
Unlike semi-supervised methods relying on a few images with mask annotations during training, methods of this group aim to learn semantic masks solely from text-based guidance. 
We roughly divide existing methods into two categories based on their cross-modal alignment between the image and text domains.

\vspace{-0.18in}
\paragraph{Image-text alignment.} These methods train an image encoder alongside a text encoder to align pairs of image and text in a joint embedding space. 
They use zero-shot transfer to enable the encoders to produce segmentation results. 
GroupViT~\cite{xu2022groupvit} introduces a bottom-up approach within Transformers, grouping image patches into regions and utilizing object semantics derived from texts to guide training. 
%
SimSeg~\cite{yi2023simple} further introduces a pretraining method that densely aligns visual and language representations, enabling the trained image encoder to generate segmentation masks in a zero-shot manner.

\vspace{-0.18in}
\paragraph{Region-text alignment.}
Another line of research targets at aligning the embedding of a region, instead of the whole image, with text descriptions.
For instance, TCL~\cite{cha2022tcl} learns to segment specific regions within an image while ensuring consistency between the segmented region and the original text. 
It enables the model to segment the relevant region described in the text.

These methods for text-supervised semantic segmentation have shown that employing vision-language models and contrastive learning on image-text pairs enables aligning visual concepts with the meaning of the whole text. 
We notice that a text is usually a mix of multiple semantic concepts, but semantic segmentation aims to discover semantically homogeneous segments. 
To address this issue, inspired by the region-word matching techniques \cite{kim2023improving, chun2021probabilistic, song2019polysemous, lee2018stacked} for cross-modal retrieval, we introduce image-text co-decomposition, where the image and the text are decomposed into image and word segments, respectively, and contrastive learning is adopted to enforce cross-modal consensus between these image and word segments.
It turns out that image-text co-decomposition results in consistent performance gains on multiple benchmarks.

\subsection{Prompt Tuning for Vision-Language Models}

Emerged from natural language processing \cite{lester2021power, li2021prefix, liu2023pre}, prompt tuning focuses on parameter-efficient adaptation of large pre-trained models to new tasks. 
In computer vision~\cite{zhou2022coop, zhou2022cocoop, khattak2023maple, zhu2023prompt}, pioneering work such as CoOp \cite{zhou2022coop, zhou2022cocoop} incorporates learnable tokens into the CLIP text encoder, enhancing the classification task performance. 
Recent studies \eg \cite{feng2022promptdet, du2022learning, Rao_2022_CVPR} leverage prompt tuning in the text modality for extending CLIP's capabilities to various applications such as detection and segmentation tasks.
Notably, prompt learning methods are also applicable to the visual domain. 
VPT~\cite{jia2022vpt} employs prompt tuning in the visual modality by inserting learnable vectors into Vision Transformers. 
Further studies \cite{liang2023open, huang2023diversity} explore tuning methods that directly incorporate learnable prompts into the input image within the RGB domain to address downstream tasks.

Drawing inspiration from the success of these methods, our method leverages the capabilities of prompt tuning on segment feature extraction in both the visual and text domains. 
Prompt learning is beneficial in this work when applying contrastive learning to the visual and textual features extracted by a vision-language model.

\section{Methodology}
\label{sec:methodology}

\begin{figure*}[t]
    \centering
    \includegraphics[width=0.95\linewidth]{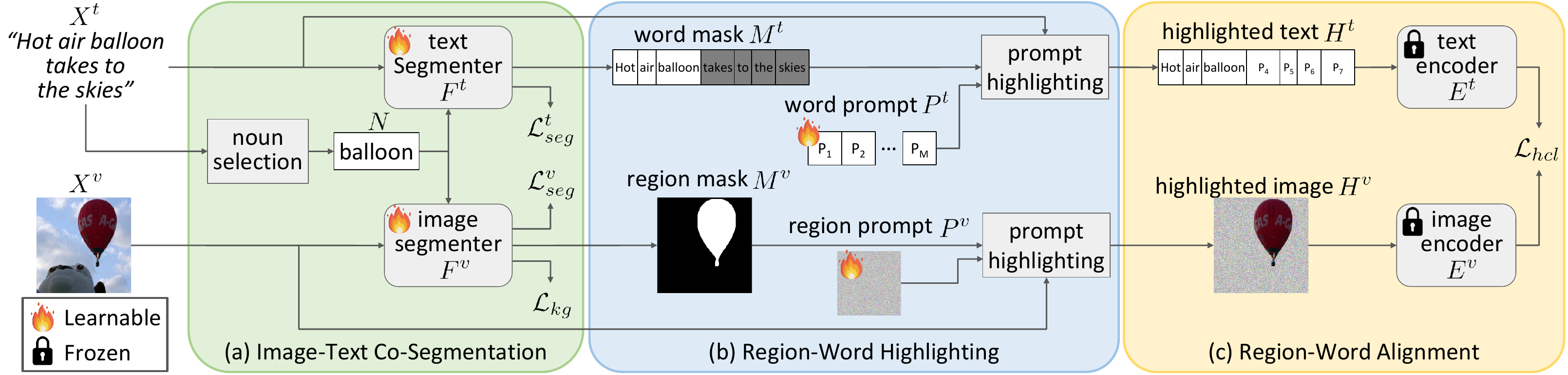}
    \vspace{-0.2cm}
    \caption{\small 
    \textbf{Training pipeline of our method for image-text co-decomposition.} 
    Our method consists of three major modules, including (a) the image-text co-segmentation module where the image and text segmenters estimate the region and word masks according to a selected noun, respectively, (b) the region-word highlighting module where the estimated masks together with two learnable prompts produce the highlighted image and text, and (c) the region-word alignment module where contrastive learning is applied to the embedded object regions and word segments to accomplish region-word alignment.
    }
    \label{fig:overallpipeline}
    \vspace{-0.3cm}
\end{figure*}

In this section, we first provide an overview of our method for image-text co-decomposition and define the notations in~\cref{overview}. 
Then, we specify the three major modules of our method, including 1) the image-text co-segmentation module in \cref{image_text_co_segmentation}, 2) the region-word highlighting module in \cref{region_word_highlighting}, and 3) the region-word alignment module in \cref{region_word_alignment}.
These modules work harmoniously to address the \ouralignment~for text-supervised semantic segmentation and enhance model performances.
Finally, implementation details are given in~\cref{implementation_details}.

\subsection{Method Overview}
\label{overview}

Image-text co-decomposition enables text-supervised segmenters to learn region-word consensus when segmenting an image $X^v$ with a paired text $X^t$.
Our method aims to jointly learn an image segmenter $F^v$ and a text segmenter $F^t$ with solely the supervision from a set of $K$ image-text pairs, $D = \{X^v_k, X^t_k\}_{k=1}^{K}$, where no annotations are given.
In addition, we optimize two learnable prompts, including a region prompt $P^v$ and a word prompt $P^t$, to alleviate the unfavorable effect of blank embeddings caused by applying a vision-language model to masked images or texts for feature extraction.

\cref{fig:overallpipeline} illustrates the pipeline of our method, consisting of three modules, including the image-text co-segmentation, region-word highlighting, and region-word alignment modules.
For an input image-text pair $(X^v, X^t)$, we initiate the process by randomly selecting a noun $N$, \eg, {\em balloon} in the figure, from the text $X^t$ using the noun selector~\cite{bird2009nltk}.
This selected noun serves as a query. 
We take the query $N$ along with the image $X^v$ as input to the image segmenter $F^v$ to generate the region mask $M^v$ showing the estimated object region specified by the query.
Similarly, a text segmenter $F^t$ takes the query $N$ and the text $X^t$ as input and estimates the word mask $M^t$ indicating the associated word segment.

Subsequently, we apply the region mask $M^v$ to the image $X^v$ to crop the estimated object region.
For the estimated background, \ie, the region outside the mask $M^v$, we crop the corresponding region from the learned region prompt $P^v$.
The highlighted image $H^v$ is yielded by combining the cropped object and background regions.
Similarly, the highlighted text $H^t$ is generated by combining the text $X^t$ inside the mask $M^t$ and the word prompt $P^t$ outside the mask $M^t$.
We extract features from the highlighted image and text by using the image encoder $E^v$ and the text encoder $E^t$ of CLIP~\cite{radford2021clip}, respectively. 
The procedure is repeated for each image-text pair and each selected noun.
It follows that the region-word alignment is accomplished by contrastive learning \cite{chen2020simclr}. 
Four loss functions, including $\mathcal{L}_\text{kg}$, $\mathcal{L}_\text{seg}^v$, $\mathcal{L}_\text{seg}^t$, and $\mathcal{L}_\text{hcl}$, are used for network optimization, and will be elaborated in the following.

\subsection{Image-Text Co-Segmentation}
\label{image_text_co_segmentation}

The image-text co-segmentation module comprises a noun selector, an image segmenter, and a text segmenter, as shown in \cref{fig:overallpipeline}a. 
Taking an image-text pair $(X^v, X^t)$ as input, this module aims at jointly identifying an object region in image $X^v$ and its accompanying word segment in text $X^t$ according to a randomly selected noun.

To begin with, we employ the noun selector~\cite{bird2009nltk}, which takes the text $X^t$ as input and extracts a set of $J$ nouns, $\{N_j\}_{j=1}^J$, in $X^t$.
For each noun $N_j$, we carry out \emph{region mask generation}, where the image segmenter $F^v$ predicts a region mask $M^v$ specifying the area in image $X^v$ relevant to noun $N_j$.
A similar task \emph{word mask generation} is performed by the text segmenter $F^t$, which seeks a word mask $M^t$ matching noun $N_j$.
The tasks of region and word mask generation are depicted as follows.

\vspace{-0.15in}
\paragraph{Region mask generation.}
The image segmenter $F^v$ takes image $X^v$ and noun $N_j$ as input. 
It encodes the image into a pixel-wise embedding $\mathbf{x}^v \in \mathbb{R}^{H \times W \times C}$, where $H \times W$ is the image resolution and $C$ is the channel dimension.
We also compute the noun embedding $\mathbf{n}_{j} \in \mathbb{R}^{C}$ for noun $N_j$.
The image segmenter generates a region mask $M^v \in \mathbb{R}^{H \times W}$ by performing the dot product between the noun embedding $\mathbf{n}_{j}$ and every location of the image embedding $\mathbf{x}^v$.

In this work, we use the image segmentation model in~\cite{cha2022tcl} to serve as the image segmenter $F^v$, and employ its corresponding loss, denoted by $\mathcal{L}_\text{seg}^v$ here, to help derive the image segmenter.
This loss considers segment regularization and contrastive learning that can be directly applied to the segmentation results along with the noun embedding. 
We use the KgCoOp method~\cite{yao2023kgcoop} to obtain the noun embedding $\mathbf{n}_{j}$, as it avoids the pitfalls of improper prompt selection. 
It appends learnable context tokens to the noun, forming pseudo-sentences for optimal prompt tuning. 
The noun embedding loss $\mathcal{L}_\text{kg}$~\cite{yao2023kgcoop} is included to improve the accuracy of these embeddings, \ie,
\begin{equation}
    \mathcal{L}_\text{kg}=||\mathbf{n}_{j}-\mathbf{n}_{j}'||^{2}_{2},
\end{equation}
where the $\mathbf{n}_{j}' \in \mathbb R^{C}$ represents the knowledge-guided noun embedding generated from hand-crafted prompts such as ``a
photo of a $N_j$'' using the text encoder.

\vspace{-0.15in}
\paragraph{Word mask generation.}
The text segmenter $F^t$ takes the text $X^t$ and the noun $N_j$ as input. 
For text feature extraction, we consider the CLIP text encoder appended with two learnable multi-head attention layers.
With the resultant feature extractor $\tilde{E}^t$, the word-wise features of text $X^t$ are obtained via $\mathbf{x}^t = \tilde{E}^t(X^t) \in \mathbb{R}^{L \times C}$, where $L$ is the text length, \ie, the number of word tokens.
The word-specific logits $\boldsymbol{\ell}_j = [\ell_{j,i}]_{i=1}^L \in \mathbb{R}^{L}$ for noun $N_j$ are computed via
\begin{equation}
\boldsymbol{\ell}_j =  w \cdot \mathbf{x}^t \mathbf{n}_j+ b,
\end{equation}
where $w$ and $b$ are two learnable parameters, and $\mathbf{n}_j \in \mathbb{R}^{C}$ is the noun embedding.

Since each word in text $X^t$ belongs to either one of the $J$ word segments associated with nouns $\{N_j\}_{j=1}^J$ or none of them, the word mask $M^t = [m^t_i]_{i=1}^L \in \mathbb{R}^{L}$ for noun $N_j$ is obtained by applying the softmax function over all the $J$ noun-associated segments, \ie,
\begin{equation}
m^{t}_{i} = \frac{\exp(\ell_{j,i})}{1 + \sum_{j'=1}^{J}{\exp(\ell_{j',i}})}, \text{ for } 1 \leq i \leq L,
\label{eq:maski}
\end{equation}
where the additional $1$ in the denominator is included for the case where word $i$ does not belong to any noun-associated segments.
The word mask $M^t$ for noun $N_j$ is produced.

According to the softmax function defined in Eq.~\ref{eq:maski}, we get the probabilities of word $i$ over $J+1$ cases, namely belonging to one of the $J$ noun-associated segments or none of them.
We compile a pseudo label vector $\mathbf{p} = \{p_i\} \in \{0,1\}^{L}$, where $p_i$ takes value $1$ if word $i$ belonging to the $j$th noun-associated segment gets the highest probability, and $0$ otherwise.
We develop the text segmentation loss $\mathcal{L}_\text{seg}^t$, which is the cross-entropy loss on the word mask $M^t$ with respect to the pseudo label vector $\mathbf{p}$, and can help learn the text segmenter $F^t$.

\subsection{Region-Word Highlighting}
\label{region_word_highlighting}

We present a prompt learning method to reliably extract features from an image region or a word segment using a vision-language model.
Specifically, we propose a region-highlighting prompt learning method and a word-highlighting prompt learning method, as shown in \cref{fig:overallpipeline}b.

\vspace{-0.15in}
\paragraph{Region highlighting prompt.}
When the region mask $M^v$ is directly applied to the image $X^v$ via $M^v * X^v$, where $*$ denotes the element-wise multiplication operation, it makes specific regions of the image being zeroed out, resulting in what we refer to as {\em blank areas}. 
When a pre-trained vision-language model like CLIP is applied to these areas, the domain distribution may shift due to the introduction of zero tokens, which are unseen in natural images. 
To mitigate this issue, we introduce a \emph{region highlighting prompt}, which is a learnable, universal image representation, denoted by $P^v$. 
This representation is used alongside the original image in the process of feature extraction. 
The highlighted image is then obtained via
\begin{equation}
    H^v = X^v * M^v + P^v * (1 - M^v).
\end{equation}
In this way, the blank areas are filled with the corresponding areas of the region prompt $P^v$ alleviating the unfavorable effect of domain shift.

\vspace{-0.15in}
\paragraph{Word highlighting prompt.}
A similar challenge arises in the text domain when applying the word mask $M^t$ to text $X^t$. 
The resultant zero tokens in the masked part unintentionally carry meanings of specific words, leading to potential inaccuracies. 
To mitigate this issue, we introduce a \emph{word highlighting prompt}, represented as a learnable, universal text representation $P^t$. 
The highlighted text $H^t$ is obtained by
\begin{equation}
    H^t = X^t * M^t + P^t * (1 - M^t).
\end{equation}
Since the masked part is filled with content from $P^t$, the risk of including unexpected text meanings is reduced.

\subsection{Region-Word Alignment}
\label{region_word_alignment}
In the following, we describe how our method achieves region-word alignment. 
Our objective is to optimize mutual evidence between the highlighted object regions and the highlighted word segments, as illustrated in \cref{fig:overallpipeline}c.

\vspace{-0.15in}
\paragraph{Contrastive loss on highlighted region-word pairs.} To achieve region-word alignment, we compute the highlighted region embedding $\mathbf{e}^v$ and highlighted word segment embedding $\mathbf{e}^t$ from the highlighted region-word pair by using the image and text encoders of CLIP by
\begin{equation}
    \mathbf{e}^v = E^v(H^v) \text{ ~ and ~ } \mathbf{e}^t = E^t(H^t),
    \label{eq:embedvt}
\end{equation}
where $E^v$ and $E^t$ are the CLIP image and text encoders, respectively.

We adopt batch optimization for model training.
Each batch has several triplets, each of which is composed of an image, its paired text, and a randomly selected noun from the text.
Each triplet yields a region embedding and a word embedding via Eq.~\ref{eq:embedvt}.
Suppose that there are $B$ triplets in this batch.
We create a similarity matrix $S = [S_{i,j}] \in \mathbb R^{B \times B}$, where $S_{i,j}$ stores the cosine similarity between the $i$th region embedding $\mathbf{e}_i^v$ and the $j$th word segment embedding $\mathbf{e}_j^t$.
We adopt the symmetric version of InfoNCE loss to develop the highlighted region-word pair contrastive loss, which enhances the similarity of related region-word pairs while reducing it for unrelated pairs:
\begin{align}
    \mathcal{L}_\text{hcl}  = &-\frac1{2B} \sum^B_{i=1} \log \frac{\exp({S}_{i,i}/\tau)}{\sum^B_{j=1} \exp({S}_{i,j}/\tau)} \nonumber \\ 
    &- \frac1{2B} \sum^B_{i=1} \log \frac{\exp({S}_{i,i}/\tau)}{\sum^B_{j=1} \exp({S}_{j,i}/\tau)},
\end{align}
where $\tau$ is a learnable temperature.
Notably, even though nouns may be selected multiple times across image-caption pairs, the corresponding highlighted regions $H^v$ and highlighted texts $H^t$ vary, ensuring the effectiveness of the InfoNCE loss in precise region-word alignment.

\vspace{-0.15in}
\paragraph{Loss functions and optimization.}
In sum, the proposed network for image-text co-decomposition is optimized using a composite loss that combines the knowledge-guided, image segmentation, text segmentation, and highlighted region-word pair contrastive losses, defined as follows:
\begin{equation}
    \mathcal{L} = 
    \lambda_\text{kg} \mathcal{L}_\text{kg} + 
    \lambda_\text{seg}^v \mathcal{L}_\text{seg}^v + 
    \lambda_\text{seg}^t \mathcal{L}_\text{seg}^t + 
    \lambda_\text{hcl} \mathcal{L}_\text{hcl}.
    \label{eq:loss}
\end{equation}

\subsection{Implementation Details}
\label{implementation_details}
We utilize NLTK's \cite{bird2009nltk} part-of-speech tagging algorithm for noun selection.
For image segmentation, we utilize TCL's image segmenter~\cite{cha2022tcl} to generate image masks, and we adopt the training loss in TCL, which relies solely on the image-caption pairs to yield $\mathcal{L}_\text{seg}^v$.
For text segmentation, we use a CLIP text encoder appended with two multi-head attention layers as the text segmenter $\tilde{E}^t$.
Our model is trained on the CC3M and CC12M datasets.
The resolution of input images is set to $224\times224$. 
For each forward pass of an image-text pair, we randomly select $2$ nouns from the text.
The loss weights are set as follows: $\lambda_\text{kg} = 8.0$, $\lambda_\text{seg}^v = 1.0$, $\lambda_\text{seg}^t = 1.0$, and $\lambda_\text{hcl} = 0.5$ in the experiments. 
We train the model with a batch size of $64$ on four NVIDIA $2080$Ti GPUs and with a learning rate of $5 \times * 10^{-6}$ for a total of $50,000$ iterations with $15,000$ warmup steps and a cosine schedule. 
AdamW optimizer~\cite{loshchilov2017decoupled} is used with a weight decay of $0.05$.
To improve the quality of the predicted mask during the evaluation phase, we adopt the post-processing approach described in TCL~\cite{cha2022tcl}, which uses pixel-adaptive mask refinement (PAMR)~\cite{araslanov2020pamr} for mask refinement.

\begin{table*}[t]
\centering
\small
\renewcommand{\arraystretch}{1.0}
\setlength{\tabcolsep}{4pt}
\begin{tabular}{l|c|c|cccccc|c}
\toprule
Methods & Publication & Dataset & {VOC}  & {Context} & {Object} & {Stuff} & {City} & {ADE} & Avg. \\
\midrule
GroupViT~\cite{xu2022groupvit} & CVPR 2022 & CC3M+CC12M+YFCC14M & 49.5 & 19.0 & 24.3 & 12.6 & 6.9 & 8.7 & 20.2\\
ViL-Seg~\cite{liu2022vilseg}             & ECCV 2022 & CC12M          & 37.3	& 18.9  & 18.1  &    &      &   &   \\
ViewCo~\cite{ren2023viewco}                                  & ICLR 2023 & CC12M+YFCC14M & 52.4 & 23.0 & 23.5 &  &    &  &   \\
OVSegmentor~\cite{xu2023learning}         & CVPR 2023 & CC12M          & 53.8 & 20.4  & 25.1  &      &    &  &   \\
SimSeg~\cite{yi2023simple}              &  CVPR 2023    & CC3M+CC12M     & \underline{57.4} & 26.2  & 29.7  &     &    &    &    \\
TCL~\cite{cha2022tcl}                &   CVPR 2023        & CC3M+CC12M     & 55.0 & \underline{30.4}  & \underline{31.6}  & \underline{22.4}  & \underline{24.0}  & \underline{17.1} & \underline{30.1} \\
SegCLIP~\cite{Luo2023SegCLIP} & ICML 2023    &   CC3M + COCO               &  52.6  &  24.7  &  26.5  &      &    &    &   \\
CoCu~\cite{xing2023rewrite}     &  NeurIPS 2023  & CC3M+CC12M+YFCC14M & 51.4& 23.6  & 22.7  & 15.2  & 22.1  & 12.3 & 24.6 \\
PGSeg~\cite{zhang2023uncovering}  & NeurIPS 2023    &  CC12M+RedCaps12M   &  53.2  &  23.8   &   28.7 &   &   &   &   \\
\midrule
\methodabbr~(Ours) & CVPR 2024 & CC3M+CC12M & \textbf{57.7} & \textbf{30.5} & \textbf{32.3} & \textbf{23.9} & \textbf{28.9} & \textbf{17.7} & \textbf{31.8} \\
\bottomrule
\end{tabular}
\vspace{-0.1cm}
\caption{\small \textbf{Text-supervised semantic segmentation performance comparison in terms of mIoU.} 
The proposed method is compared with nine SOTA methods on six popular semantic segmentation datasets: PASCAL VOC (VOC), PASCAL Context (Context), COCO-Object (Object), COCO-Stuff (Stuff), Cityscapes (City) and ADE20K (ADE). For each compared method, the dataset column lists its training datasets. Several methods used datasets in addition to CC3M and CC12M, such as YFCC14M, COCO and RedCaps12M. When applicable, we also provide an average mIoU across all six datasets. For each dataset, the best method is indicated by bold fonts, whereas the second best method is underlined. 
}
\label{table:main}
\vspace{-0.3cm}
\end{table*}

\section{Experiments}
\label{sec:experimental_results}

\subsection{Datasets and Evaluation Settings}
We utilize image-text datasets to train our proposed model and perform extensive experiments on six commonly used semantic segmentation benchmarks to validate our method.
\\
\textbf{Training datasets.} 
We trained our model on two image-text datasets, Conceptual Captions 3M (CC3M) \cite{sharma2018cc3m} and Conceptual 12M (CC12M) \cite{changpinyo2021cc12m} containing 3M and 12M image-text pairs respectively. They have been widely adopted for training text-supervised semantic segmentation methods. 
\\
\textbf{Evaluation datasets.} 
We used six zero-shot semantic segmentation benchmarks to validate the zero-shot transfer capability of our model on categories that were not specifically trained.
As in previous work~\cite{cha2022tcl}, the benchmarks can be categorized into two groups, with and without background classes.
Benchmarks with a background generally label areas that do not belong to any predefined categories as ``background,'' which is usually removed by considering a probability threshold in text-supervised semantic segmentation. For this category, we use the validation split of the following datasets: PASCAL VOC 2012~\cite{everingham2010pascal}, PASCAL Context~\cite{mottaghi2014context59}, and COCO-Object~\cite{caesar2018cocostuff}. They each contain 20, 59, and 80 foreground classes, respectively, with an additional background class.
For the ``without background category,'' we evaluated our model with the validation split of COCO-Stuff \cite{caesar2018cocostuff}, Cityscapes \cite{cordts2016cityscapes}, and ADE20K \cite{zhou2019ade20k} datasets. Each of them contains 171, 19, and 150 classes, respectively.
In this category, all images are fully annotated, which is exceptionally challenging.
Using datasets in this category, our model can be tested for its ability to recognize a variety of concepts.
We employ mean intersection-over-union (mIoU) as our evaluation metric.

For zero-shot semantic segmentation evaluation, we rely solely on the image segmenter. The image segmenter processes the input image in conjunction with class names from each dataset to produce segmentation predictions. In accordance with the settings of prior work~\cite{cha2022tcl}, we adopt the class names provided by the default version of MMSegmentation \cite{mmseg2020} and adhere to its post-processing methodology.

\subsection{Quantitative Comparisons}

We compare the proposed method with nine text-supervised semantic segmentation methods on the six datasets. 
\cref{table:main} reports the mIoU values. The numbers have been taken directly from the original papers. 
All methods were tested on the three datasets of the ``with background class,'' but only three methods (GroupViT~\cite{xu2022groupvit}, CoCu~\cite{xing2023rewrite} and TCL~\cite{cha2022tcl}) were tested on the dataset of the ``without background class.'' 
For those three methods, we also report their average mIoU values across all six datasets. 
It is also worth noting that these methods use different combinations of training datasets, as indicated in the dataset column of \cref{table:main}.

Our method achieves the best performance in all six datasets, while TCL~\cite{cha2022tcl} and SimSeg~\cite{yi2023simple} are the runners-up. 
In terms of average mIoU, our method~(\methodabbr) achieves 31.8 whereas TCL achieves 30.1, resulting in a 5.65\% improvement. 
The result demonstrates the effectiveness of our image-text co-decomposition method in addressing the alignment-level train-test discrepancy that exists in previous methods by directly learning the region-word alignment.

\subsection{Qualitative Results}

\paragraph{Visual comparison with existing methods.}

\begin{figure*}[t]
    \centering
    \vspace{-0.2cm}
    \includegraphics[width=0.9\linewidth]{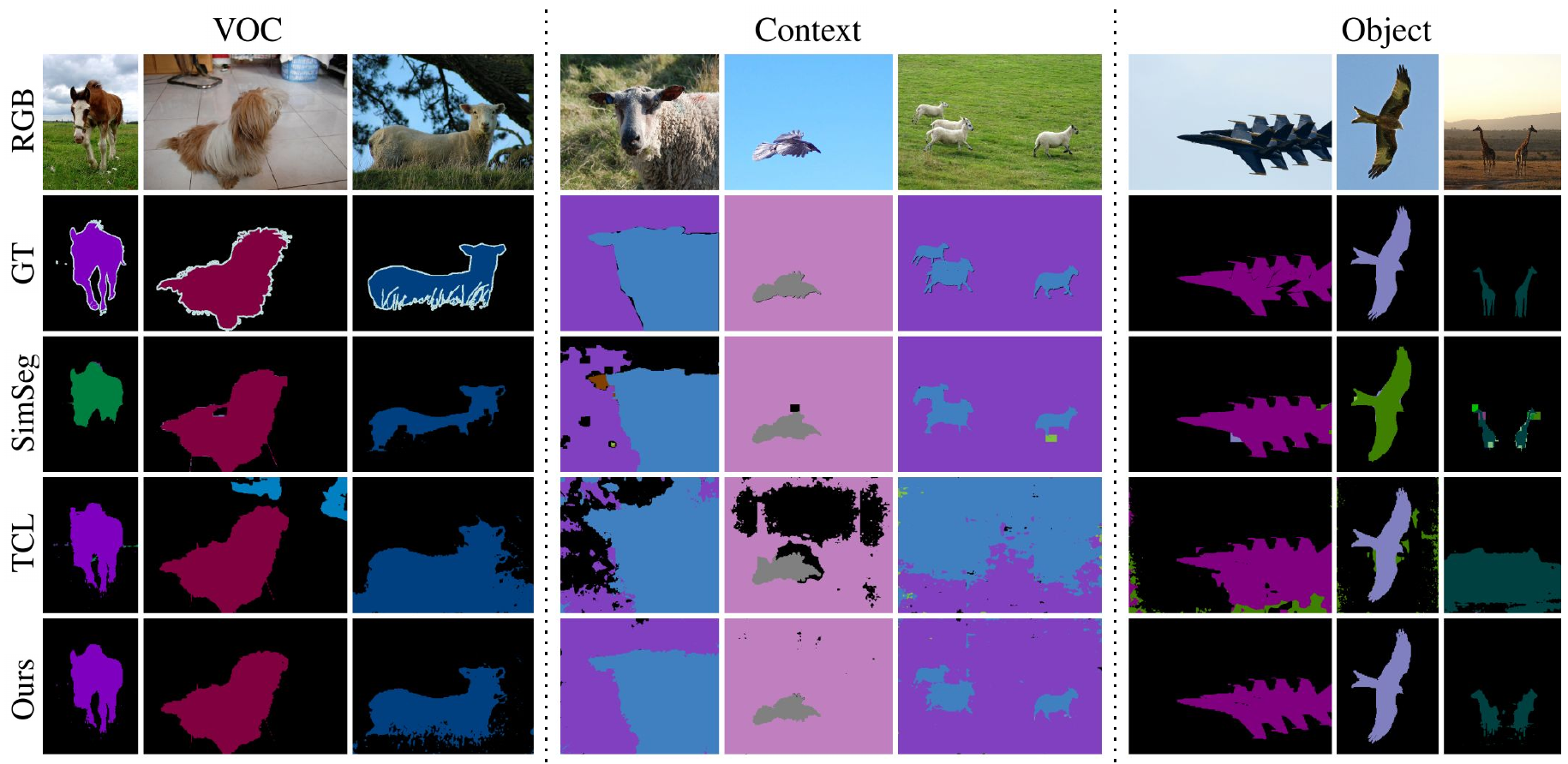}
    \caption{\textbf{Qualitative comparisons.}
    The proposed method is compared with the two most competitive methods, TCL~\cite{cha2022tcl} and SimSeg~\cite{yi2023simple}, on PASCAL VOC, PASCAL Context, and COCO Object datasets. 
    Our method provides more precise object boundaries and effectively localizes objects within images without misclassification, leading to more accurate segmentation.
    }
    \label{fig:sota-visualization}
\vspace{-0.3cm}
\end{figure*}

\cref{fig:sota-visualization} visually compares the segmentation results of our methods and two runners-up, TCL~\cite{cha2022tcl} and SimSeg~\cite{yi2023simple}, on the PASCAL VOC, PASCAL Context, and COCO Object datasets.

This figure illustrates the fundamental benefit of our approach, which involves the direct learning of region-word alignments. Our model effectively establishes a strong connection between object regions and word segments, allowing a better understanding of how objects are represented in images. Through this enhanced understanding, both segmentation quality and localization capabilities can be improved.
As a result, our method provides more accurate classification and more precise masks than other methods.

The SimSeg\cite{yi2023simple} model, which learns from image-text alignments, occasionally assigns objects to the wrong classes. On the other hand, TCL~\cite{cha2022tcl}, which is based on region-text alignment, produces coarser semantic masks. Accordingly, these observed limitations are most likely a result of the alignment-level discrepancy between the train and test, which may lead to suboptimal performance.

\paragraph{Visualization of image-text co-segmentation results.}
    
\begin{figure*}[t]
    \centering
    \vspace{-0.3cm}
    \includegraphics[width=\linewidth]{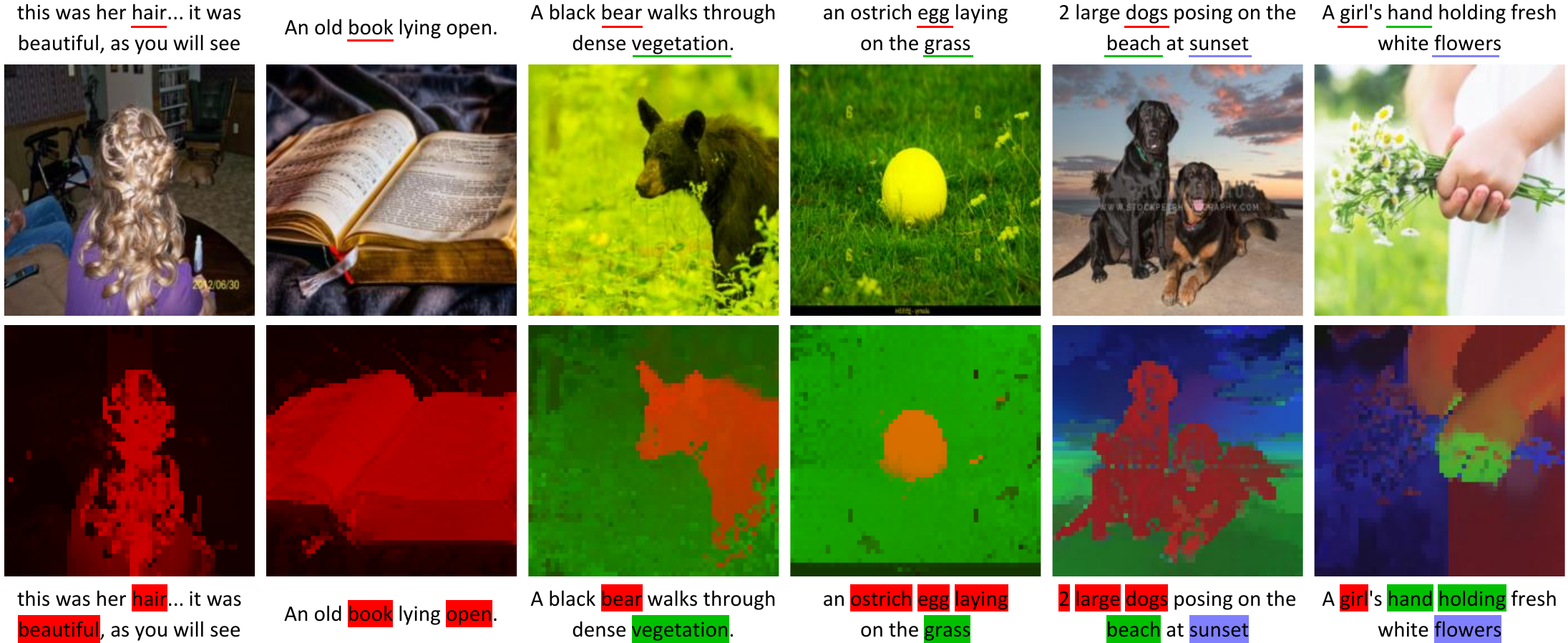}
    \vspace{-0.3cm}
    \caption{\small \textbf{Visualization of the results of our image-text co-decomposition method.} 
    The first two rows display text and images, representing input image-text pairs. In each text, nouns are underlined with different colors. Our method uses these nouns as queries for performing image-text co-decomposition. Using our image-text co-decomposition method, the last two rows depict the method's output, where regions and word segments associated with different nouns appear in corresponding colors.
    }
    \vspace{-0.1cm}
    \label{fig:codec}
\end{figure*}

\cref{fig:codec} presents a visualization of the results obtained by our model. We denote regions and word segments associated with the different nouns in the corresponding colors. It demonstrates that our method effectively segments object regions within images based on various input nouns. It simultaneously segments corresponding word segments within the associated text, creating a harmonious alignment between the object region and the word segment.

The region-word alignment plays a pivotal role in our approach, serving as a supervisory signal for the model. By taking advantage of this alignment, our model not only performs visual localization but also captures correlations within the language domain. It indicates that our trained model possesses a more comprehensive understanding of the segmentation task.

\subsection{Ablation Study}

\begin{table}[t]
\centering
\renewcommand\arraystretch{1.2} 
\renewcommand\tabcolsep{6pt} 
\resizebox{\columnwidth}{!}{%
\begin{tabular}{ccc|ccccccc}
\toprule
C. & W. & R. & {VOC}  & {Context} & {Object} & {Stuff} & {City} & {ADE} & Avg. \\
\midrule
           &            &            & 54.4 & 27.6 & 32.7 & 22.5 & 25.0 & 16.6 & 29.8 \\
\checkmark &            &            & 56.2 & 29.2 & \textbf{32.9} & 23.3 & 27.5 & 17.0 & 31.0 \\
\checkmark & \checkmark &            & 56.1 & 29.3 & 32.6 & 23.6 & \textbf{29.0} & 17.3 & 31.3 \\
\checkmark & \checkmark & \checkmark & \textbf{57.7} & \textbf{30.5} & 32.3 & \textbf{23.9} & 28.9 & \textbf{17.7} & \textbf{31.8} \\
\bottomrule
\end{tabular}
}
\caption{\textbf{Ablation study.} The baseline model is augmented with the image-text co-decomposition method (C.), the word highlighting prompt (W.), and the region highlighting prompt (R.), one at a time. We report the mIoU values of the resultant models on the six datasets and their averages.
}
\label{tab:ablation}
\vspace{-0.5cm}
\end{table}

\paragraph{Contributions of individual components.}
The ablation study in 
\cref{tab:ablation} assesses the contribution of the proposed components, including the image-text co-decomposition method, the word highlighting prompt, and the region highlighting prompt. 
Without the co-decomposition method, our baseline model only trains the image segmenter, resulting in an average mIoU of 29.8.
Afterward, each proposed component is added to the baseline model one at a time to verify its contribution.
As a result of adding the image-text co-decomposition module alone, the average mIoU has been increased to 31.0.
It suggests that the image-text co-decomposition method can achieve region-word alignment and enhance localization capability.
The model is further enhanced with the addition of word highlighting prompts and image highlighting prompts, resulting in further performance improvement.  
It demonstrates that the highlighting prompt learning method enhances feature extraction and strengthens alignment between regions and words.

\begin{table}[t]
\centering
\renewcommand\arraystretch{1.2} 
\renewcommand\tabcolsep{6pt} 
\resizebox{0.7\columnwidth}{!}{%
\begin{tabular}{@{}c|cccccc@{}} 
\toprule
    $\lambda_\text{hcl}$ & 0.05 & 0.1 & 0.25 & 0.5 & 0.75 & 1.0 \\
    \midrule
    Avg.   & 30.6 & 31.2 & 31.7 & \textbf{31.8} & 31.5 & 30.8 \\
    \bottomrule
\end{tabular}
}\caption{\textbf{Sensitivity analysis on the hyperparameter $\lambda_\text{hcl} $.} By varying $\lambda_\text{hcl}$, we examine the corresponding average mIoU values of all six datasets.}
\label{tab:hyperparam}
\end{table}

\paragraph{Hyperparameter sensitivity analysis.}
\cref{tab:hyperparam} investigates the impact of the loss weight for the highlighted region-word pair contrastive loss, denoted as $\lambda_\text{hcl}$ in \cref{eq:loss}.
We observe that, when we apply the highlighted region-word pair contrastive loss in our training phase, the performance consistently outperforms our baseline model. 
The method is robust to the parameter to some degree as it achieves reasonable performance for a wide range of values.
When $\lambda_\text{hcl}$ is set to 0.5, our model achieves a peak score of 31.8. 
It is evident from these results that the image-text co-decomposition method is superior to the image-text decomposition method for achieving region-word alignment. 

\paragraph{Effectiveness of jointly decomposing text.}
We validate the effectiveness of decomposing text by assessing the performance enhancement achieved by generating word masks, as opposed to simply using extracted nouns.
This experiment is conducted by modifying the calculation of $\mathcal{L}_\text{hcl}$.
Instead of using word segment embeddings as mentioned in~\cref{region_word_alignment}, we opt to compute the similarity matrix $S$ using region embeddings with the embeddings of \textit{individual nouns}. The average mIoU across all benchmarks is $30.2\%$, which is below our method's $31.8\%$.
This indicates the benefits of using word segments encompassing extra words associated with each noun. The contextual information encoded in these additional words can serve as valuable supervisory signals, thereby improving performance.


\section{Conclusions}
\label{sec:conclusions}

We propose \methodfull~(\methodabbr) to address cross-domain alignment discrepancies in the existing methods for text-supervised semantic segmentation.
First, our method decomposes image-text pairs into corresponding regions and word segments to enforce the region-word alignment.
\methodabbr, underpinned by contrastive learning, alleviates the train-test discrepancy by unifying image-text and region-text alignments to region-word alignment.
Then, we introduce a region-highlighting prompt learning method to enhance feature extraction on masked images or texts for precise region-word alignment.
Moreover, \methodabbr~surpasses state-of-the-art methods in zero-shot semantic segmentation across six benchmark datasets.
This novel approach opens new possibilities for research in vision-language models and their broader applications in computer vision.

\section{Acknowledgement}
\label{sec:acknowledgement}

This work was supported in part by the National Science and Technology Council (NSTC) under grants 112-2221-E-A49-090-MY3, 111-2628-E-A49-025-MY3, 112-2634-F-002-005, 112-2634-F-002-006, and 110-2221-E-002-124-MY3, and NTU under grants 112L9009. This work was funded in part by MediaTek and NVIDIA.

{
    \small
    \bibliographystyle{ieeenat_fullname}
    \bibliography{main}
}

\clearpage
\setcounter{page}{1}
\maketitlesupplementary

\section{Additional Qualitative Results}
\label{sec:additionalqualitativeresults}

\subsection{Ablation study visualization}

\begin{figure}[t]
    \centering
    \includegraphics[width=\columnwidth]{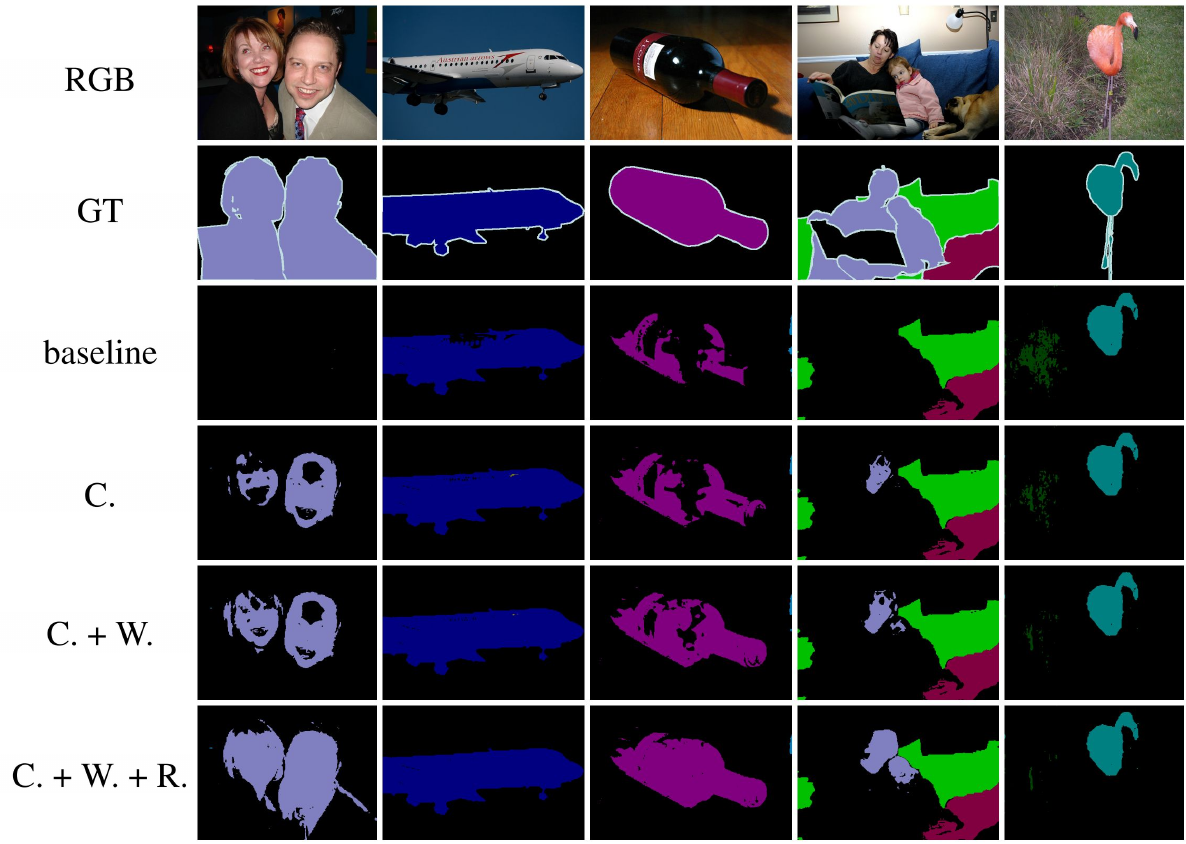}
    \vspace{-0.5cm}
    \caption{\textbf{Ablation studies.} We improve the baseline model by incrementally including (C.) the image-text co-decomposition module, (W.) the word highlighting prompt, and (R.) the region highlighting prompt. 
    We present the segmentation results of the resulting models on the images of the PASCAL VOC~\cite{everingham2010pascal} dataset.}
    \vspace{-0.2cm}
    \label{fig:ablation-visualization}
\end{figure}

In the following, we conduct ablation studies by visualizing the effects of the proposed components in our method, including the image-text co-decomposition method, the word highlighting prompt, and the region highlighting prompt.
To this end, \cref{fig:ablation-visualization} offers the visual comparison of segmentation results produced by the variants of our method on five images of the PASCAL VOC~\cite{everingham2010pascal} dataset.


The image-text co-decomposition module equips the model with the region-word alignment ability to localize objects in the images accurately. This module aligns words with corresponding regions in the image, leading to more precise segmentation results. Furthermore, both the word and region highlighting prompts contribute to feature extraction, improving the model's ability to capture the details of the objects.
Hence, the resultant model is more effective in segmenting the whole objects of interest.

\begin{figure}[h]
    \centering
    \includegraphics[width=\columnwidth]{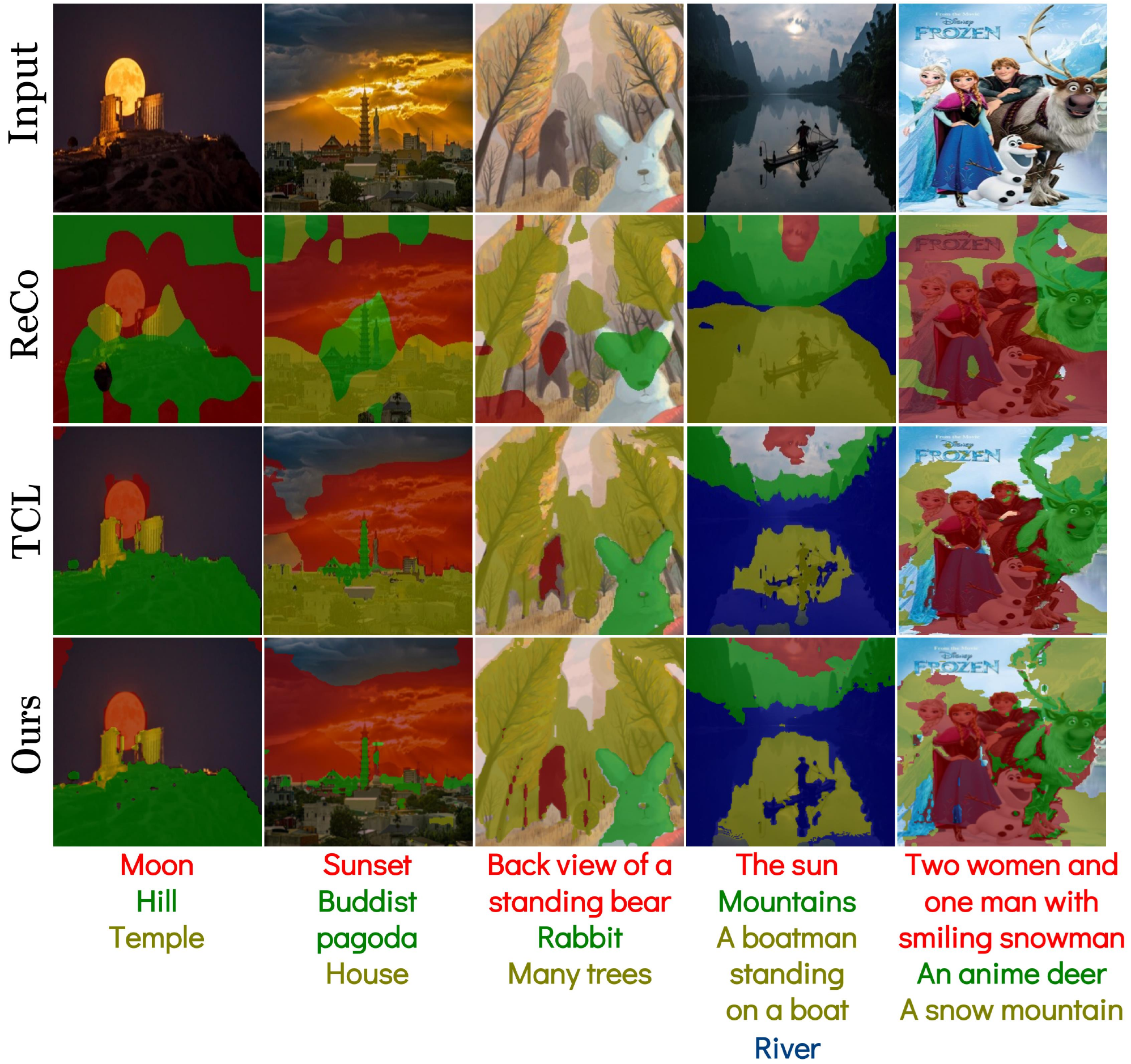}
    \vspace{-0.5cm}
    \caption{\textbf{Examples in the wild.} We show predictions on wild images with free-form text queries. Texts used as target classes are shown at the bottom of the images.}
    \vspace{-0.2cm}
    \label{fig:wild-examples}
\end{figure}

\subsection{Multi-noun queries}
Fig.~\ref{fig:wild-examples} shows predictions on wild web images with various text queries using the same images and queries selected from Fig. 5 of TCL~\cite{cha2022tcl}. 
Although our method is primarily designed and trained for single-noun queries, the figure demonstrates its effectiveness in processing more complex queries.

\subsection{Failure case visualization}

\begin{figure}[t]
    \centering
    \includegraphics[width=\columnwidth]{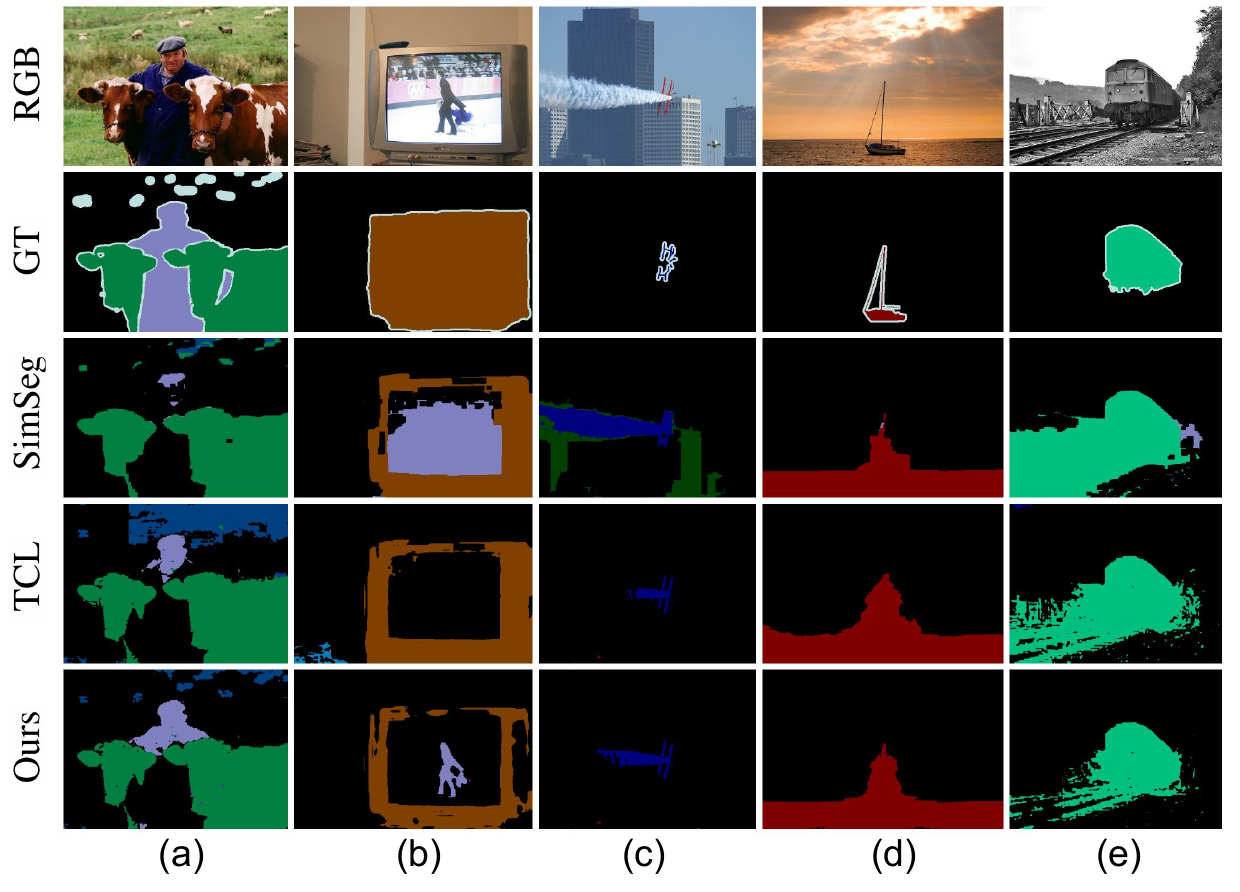}
    \vspace{-0.5cm}
    \caption{\textbf{Failure cases.}
    The proposed method is compared with the two most competitive methods, TCL~\cite{cha2022tcl} and SimSeg~\cite{yi2023simple}, on the images of the PASCAL VOC~\cite{everingham2010pascal} dataset. 
    }
    \vspace{-0.2cm}
    \label{fig:failure-case-visualization}
\end{figure}

In \cref{fig:failure-case-visualization}, we show several failure cases of our method and two competing methods, TCL~\cite{cha2022tcl} and SimSeg~\cite{yi2023simple}, on the images of the PASCAL VOC~\cite{everingham2010pascal} dataset.

The first example in \cref{fig:failure-case-visualization}a shows a common limitation of existing methods: When segmenting the ``person'' class, most methods focus on the most distinctive areas, namely the face in this example, and suffer from the variations in the clothes, resulting in the segment that does not cover the entire person.
The second example in \cref{fig:failure-case-visualization}b depicts a scenario, where unexpected variations are present, \ie, people showing in a television monitor.
All three methods segment the outer borders of the monitor.
Compared to TCL and SimSeg, our method can further segment the individuals within the monitor.
Although the ground truth covers the entire TV monitor, this example validates the effectiveness of our model in localizing the individuals present on the screen.

\cref{fig:failure-case-visualization}c, \cref{fig:failure-case-visualization}d, and \cref{fig:failure-case-visualization}e showcase instances where co-occurrent objects, such as trains and tracks, airplanes and contrails, and boats and water, tend to be segmented together even though they are of different semantic categories. 
This is a challenge for our method and the two competing methods TCL~\cite{cha2022tcl} and SimSeg~\cite{yi2023simple}.
These visualization examples emphasize the difficulties of accurate segmentation and the challenges in aligning model predictions with ground truth annotations. 
They provide insights into the limitations of current segmentation approaches and suggest future research directions.

\section{More Implementation Details}
\paragraph{Training time.}
On four NVIDIA 2080Ti GPUs, it takes eight hours to train the baseline model with only the image segmenter. On the same devices, it takes twelve hours to train our image-text co-decomposition method, which requires training an additional text segmenter. In light of the improved performance as described above, the longer training period can be justified.

\end{document}